# Design analysis of an innovative parallel robot for minimally invasive pancreatic surgery

Doina Pisla[1,2][0000-0001-7014-9431], Alexandru Pusca[1*][0000-0002-5804-575X], Andrei Caprariu[1][0009-0008-1152-8077], Adrian Pisla[1][0000-0002-5531-6913], Bogdan Gherman[1][0000-0002-4427-6231], Calin Vaida[1][0000-0003-2822-9790], Damien Chablat[1,3][0000-0001-7847-6162]

[1] CESTER, Technical University of Cluj-Napoca, Memorandumului 28, 400114 Cluj-Napoca, Romania
[2] Technical Sciences Academy of Romania, B-dul Dacia, 26, 030167 Bucharest, Romania
[3] École Centrale Nantes, Nantes Université, CNRS, LS2N, UMR 6004, F-44000 Nantes, France
*Corresponding author: `Alexandru.Pusca@mep.utcluj.ro`

**Abstract.** This paper focuses on the design of a parallel robot designed for robotic assisted minimally invasive pancreatic surgery. Two alternative architectures, called ATHENA-1 and ATHENA-2, each with 4 degrees of freedom (DOF) are proposed. Their kinematic schemes are presented, and the conceptual 3D CAD models are illustrated. Based on these, two Finite Element Method (FEM) simulations were performed to determine which architecture has the higher stiffness. A workspace quantitative analysis is performed to further assess the usability of the two proposed parallel architectures related to the medical tasks. The obtained results are used to select the architecture which fit the required design criteria and will be used to develop the experimental model of the surgical robot.

## 1 Introduction

Pancreatic cancer has the second highest incidence rate of new cases in Europe, with countries such as Hungary, Slovakia, Czech Republic and Serbia having the highest rates [1]. According to Wong's research [2], the incidence of pancreatic cancer is more common in developed countries compared to developing countries, with the lowest incidence rates reported in Africa and Central and South Asia. Regarding the history and evolution of pancreatic cancer, tumor resection remains the most effective way to reduce the mortality rate. Studies [3-6] present advances in pancreatic surgery, considered the most complex surgery in the abdominal cavity compared to other types of surgery, such as cholecystectomy, bowel resection or bariatric procedures. This is due to anatomical complexity, limited working space, difficult exposure, oncological considerations and fragility of the tissues that compose or border the pancreas. Currently, pancreatic cancer surgery has become much safer due to technological advances, types of therapies and the integration of robots [7-10] in performing this procedure, with the associated mortality rate under 3% for all three main types of surgery (Pancreaticoduodenectomy - PD, Distal Pancreatectomy - DP and Total Pancreatectomy - TP) [11-13]. Due to the high complexity of pancreatic surgery, several robots, such as those developed by the companies Intuitive Surgical (da Vinci, da Vinci S, da Vinci Si, da Vinci Xi and da Vinci SP), Asensus Surgical (Senhance robot) or CMR Surgical (Versius robot), have been developed and used in pancreatic surgery in order to overcome challenges associated with open surgery (OS) or laparoscopic surgery, such as major reconstructions required after pancreatic resection [11-12], size of incision, blood loss, long recovery time, postoperative trauma, hand tremor or anastomosis required for the reconstruction phase [13-16]. In addition to these advantages of the robots currently used in pancreatic surgery, the major disadvantages still present in this field are reflected in the high cost of these robots, the volume occupied into the operating room, the configuration of the operating room, the learning curve of the surgeon, the lack of tactile feedback, the triangulation effect of the instruments, the limited intraoperative field, the limited workspace, the complexity of the reconstruction phase and the simultaneous handling of a maximum of two active instruments using a master-slave architecture [17]. Due to the disadvantages presented previously and the problems which persist in pancreatic surgery, research to develop innovative robotic solutions remains open. Stiffness of the current systems used in minimally invasive surgery according to the data presented in [18-19] is between 5-10 N/mm and the maximum displacement is between 0.1 and 1.4 millimeters depending on the configurations used. Thus, this paper presents a stiffness analysis of the two robots proposed for minimally invasive pancreatic surgery, performed using the FEM, the final results contributing to the selection of the final robot to be developed in order to provide the surgeon with the possibility to precisely manipulate a third active instrument used in the dissection (active instrument used as a retractor) and reconstruction stages (atraumatic active instrument used for the pancreas reconstruction).

Following the introduction, the paper is structured as follows: Section II presents the kinematic schemes and input-output equations of the two proposed parallel robots developed for minimally invasive pancreatic surgery. Section III analyzes these robots using FEM applied to their 3D designs. Section IV presents the workspace analysis of robots, and the final section provides a series of conclusions and future developments.

## 2 The kinematic architecture of the surgical parallel robot

This section presents the kinematic schemes and input-output equations of two 4-degrees of freedom (DOF) parallel robots architectures, ATHENA-1 and ATHENA-2 [20] developed for minimally invasive pancreatic surgery. ATHENA-1 (Fig. 1) consists of three active prismatic joints $q_i, i=1...3$ (red) and one active revolute joint $q_4$, three passive prismatic joints $P_{iR}, i=1...3$, two passive universal joints $U_{iR}, i=1...2$ and nine passive revolute joints $R_{iR}, i=1...9$. The geometrical parameters of the ATHENA-1 parallel architecture are the lengths of the elements used to connect the joints denoted by $l_i, i=1...5$, $l_{2min}$ (min. stroke for $q_3$), $l$ as the length of the active instrument, $l_{ins}$ as the insertion length of the active instrument measured from the insertion point-the Remote

Center of Motion (RCM) [21] and $l_{0i}$, $i=1...3$ are the lengths between the distance from the *OXYZ* coordinate system of the robot relative to the fixed coordinates system with origin in the RCM of the spherical mechanism ($O_{RCM}$, $X_{RCM}$, $Y_{RCM}$, $Z_{RCM}$). The parallel spherical mechanism used to ensure the RCM of the active instrument consists of five passive revolute joints $R_{Si}$, $i=1...5$ and one passive cylindrical joint $C_{S1}$, which provide instrument rotation ($\psi$, $\theta$) and insertion, while tip rotation ($\varphi$) is achieved directly by $q_4$. The connection between parallel robot and the spherical mechanism is achieved using two spherical joints $S_i$, $i=1...2$, which allow the additional orientation and positioning of the spherical mechanism during the preplanning stage (after the orientation and additional position of the spherical mechanism are established, the spherical joints are locked).

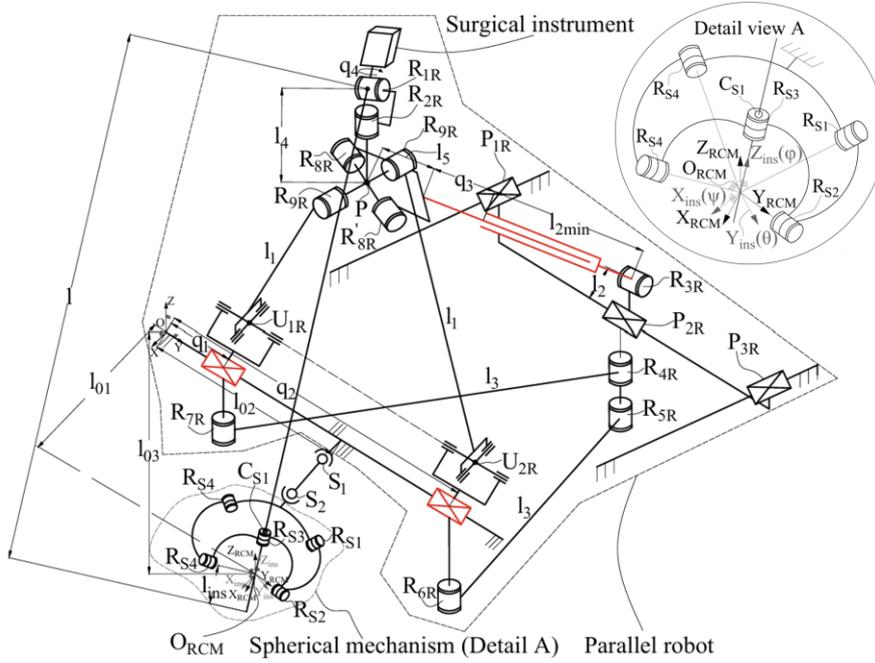

**Fig. 1.** Kinematic scheme of the ATHENA-1 parallel architecture

According to the kinematic scheme of the ATHENA-1 (Fig. 1) and the data presented in [22] the input-output equation for the active joints ($q_i$, $i=1...4$) displacements can be expressed using Eq. 1-5.

$$f_1 : l_{02} + (q_1 + (q_2/2)) - Y_P = 0 \tag{1}$$

$$f_2 : \left(l_4 + \sqrt{l_1^2 - (q_2 - (q_1/2))^2}\right)^2 - (Z_P - l_{03})^2 - (X_P - l_{01})^2 = 0 \tag{2}$$

$$f_3 : (q_3 + l_{2\min} + l_5)^2 - \left(X_P - l_{01} - l_4 \cos(\lambda) + \sqrt{l_3^2 - (q_2 - (q_1/2))^2}\right)^2 - (Z_P - l_4 \sin(\lambda) - l_{03})^2 = 0 \tag{3}$$

$$f_4 : \sin q_4 - \sin \varphi = 0 \tag{4}$$

where $X_P$, $Y_P$, $Z_P$ and $\lambda$ are:

$$X_p = \cos(\psi)\sin(\theta)(l - l_{ins}), Y_p = \sin(\psi)\sin(\theta)(l - l_{ins})$$
$$Z_p = \cos(\theta)(l - l_{ins}), \lambda = \operatorname{atan2}(Z_P - l_{03}, X_P - l_{01}) \quad (5)$$

The kinematic scheme of the ATHENA-2 parallel architecture is illustrated in Fig. 2. The green kinematic chain has replaced the active prismatic joint used for the ATHENA-1 parallel architecture. It consists of the active revolute joint ($q_3$), the passive revolute joints $R_{3R}$, and $R_{8R}$, and the two $l_1$ links. All other kinematic chains and geometrical parameters remain identical with the ones of ATHENA-1.

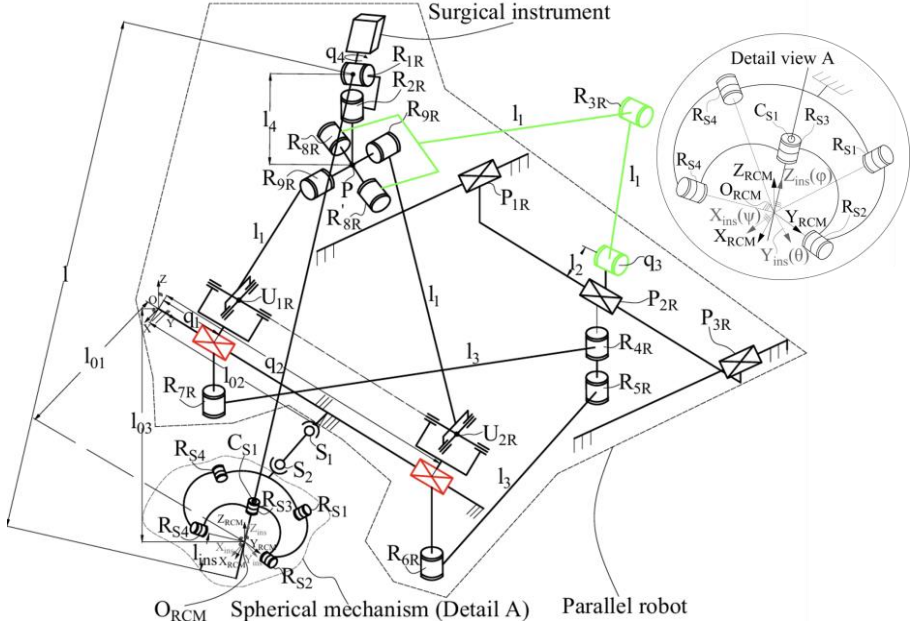

**Fig. 2.** Kinematic scheme of the ATHENA-2 parallel architecture

Based on the kinematic scheme illustrated in Fig. 2, as well as the detailed study presented in [23], the input-output equations for the ATHENA-2 parallel robot are expressed as follows:

$$f_1 : l_{02} + (q_1 + (q_2/2)) - Y_P = 0 \quad (6)$$

$$f_2 : \left(l_4 + \sqrt{l_1^2 - (q_2 - (q_1/2))^2}\right)^2 - (Z_P - l_{03})^2 - (X_P - l_{01})^2 = 0 \quad (7)$$

$$f_3 : ((t_1 - l_4) \cdot \sin(t_3) + t_2 + (t_1 - l_4) \cdot \cos(t_3) - l_2 \cdot \sin(q_3))^2 - l_2^2 = 0 \quad (8)$$

$$f_4 : \sin q_4 - \sin \varphi = 0 \quad (9)$$

where:

$$t_1 = (\sqrt{(X_p + l_0)^2 + Z_P^2} - l_4)^2, \; t_2 = \sqrt{(l_3^2 - ((q_2 - q_1)/2)^2) - l_2 \cdot \cos(q_3))^2},$$
$$t_3 = \operatorname{atan2}(((X_P + l_0) \cdot ((X_P + l_0)^2 + Z_P^2))^{(-1/2)}), Z_P \cdot ((X_P + l_0)^2 + Z_P^2)^{(-1/2)}) \quad (10)$$

## 3      Stiffness analysis of ATHENA-1 and ATHENA-2

Based on the previously presented kinematic schemes, the preliminary CAD models of the ATHENA-1 and ATHENA-2 parallel architectures was generated using Siemens NX. To validate the design and to ensure that the intended functionality of the two parallel robots can be achieved, kinematic simulations were performed in Siemens NX Motion [22-23]. The structural performance of ATHENA-1 and ATHENA-2 was evaluated by the Finite Element Method (FEM) simulations. For each alternative design (ATHENA-1 and ATHENA-2), static simulations were conducted in Abaqus Standard to assess the overall stiffness and to identify the structural components that are subjected to high stresses or deformations. Given the importance of stiffness as a design criterion for surgical robots, the results of the FEM simulations were essential for deciding which design will be chosen.

Figure 3 presents the 3D design of ATHENA-1 parallel architecture. The robot is driven by three linear active lead-screw actuators: $q_1$, $q_2$, and $q_3$. The rotation of the active instrument through the angle $\psi$ is provided by translation movement along the $OY$ axis, performed by $q_1$ and $q_2$ which are also driving the two passive universal joints ($U_{1R}$ and $U_{2R}$). The passive universal joints $U_{1R}$ and $U_{2R}$ allow the orientation ($\theta$) of the active instrument through the translation movement generated by the third active linear actuator $q_3$ which of a linear piston fixed on an axis that allows free translations along the $OX$ and the $OY$ axes using linear sliders translating along rails mounted on the robot frame. In addition to these elements, rods and passive revolute joints are used to provide the transmission of movement from the three linear actuators to the active instrument. To provide the RCM of the active instrument mounted on the ATHENA-1 parallel architecture, a spherical parallel mechanism consisting of 5 arms connected to each other was designed, this passive mechanism providing three orientations ($\psi$, $\theta$, $\varphi$) and one insertion of the active instrument ($l_{ins}$).

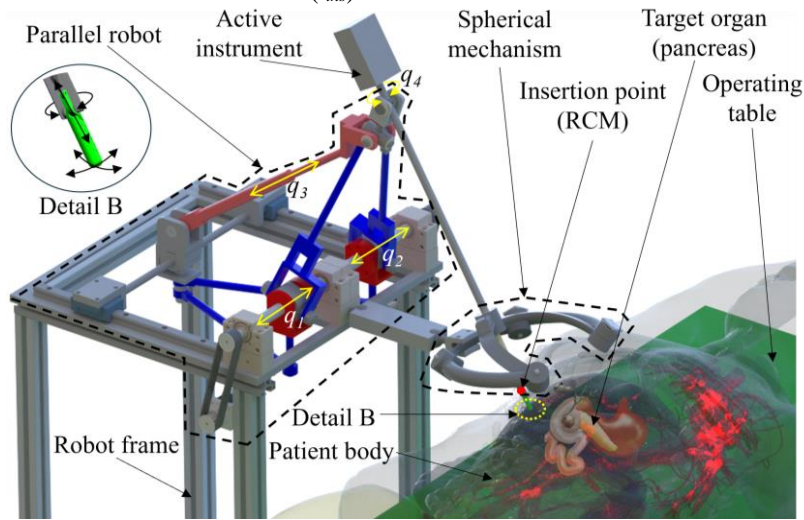

**Fig. 3.** The 3D model of ATHENA-1

The 3D design of the ATHENA-2 parallel architecture is illustrated in figure 4, where the linear piston has been replaced by an active revolute actuator ($q_3$). Based on the previously described 3D models, the FEM models of the two parallel robots were generated and static simulations were performed for each robot. The purpose of the static FEM simulations was to determine the overall stiffness and to provide insights regarding the mechanical behavior of components subjected to high stresses and deformations for each individual robot.

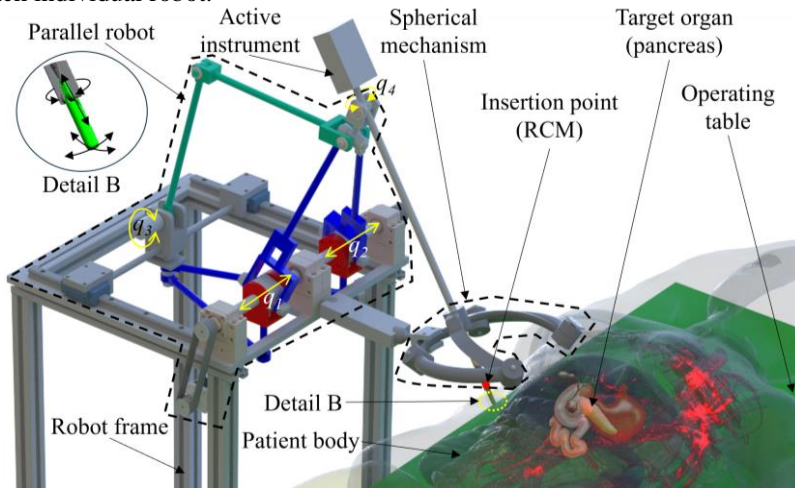

**Fig. 4.** The 3D model of ATHENA-2

Fig. 5a presents the load case scenario of the static simulation that was performed for ATHENA-1. First, using Siemens NX the 3D model of the robot was positioned and oriented in the configuration where the instrument was completely inserted into the patient's body. The described position is achieved at the limit of the active actuators' strokes. After achieving the desired configuration, finite elements have been generated for all the relevant components of the robot, the materials have been assigned to each mechanical element and the connections and interactions between the components of the robot have been defined. The active joints were modeled using kinematic coupling constraints which are locking the active joints in the position where the robot is most loaded, while for the passive joints, distributing coupling and hinge/cylindrical connector elements have been used. As presented in Figure 5a, fixed boundary conditions are applied to the components that are directly or indirectly fixed on the frame of the robot. A concentrated force (F) with a magnitude of 30 N was applied along the axis of the active instrument while an additional gravitational load (magnitude = 1g) was applied for all the components of the robot. A static nonlinear step was defined, and the simulation was computed.

Following the same methodology, the FEM model of ATHENA-2 parallel architecture was generated, and the simulation was computed. The load case scenario of the simulation that is presented in Figure 5b. For consistent results, a similar position was used (in the extended position), the same materials were assigned, the connections/interactions between the components were defined in the same way, the same boundary conditions were defined, and the same loads were applied. The results of the two FEM performed for ATHENA-1 and ATHENA-2 parallel architectures are presented in Fig. 6-7, Fig. 6a and Fig. 6b present the overall displacement while Fig. 7a and Fig. 7b present the von Mises stress distribution after applying a concentrated force of 30 N along the active instrument, highlighting the areas with the maximum stresses.

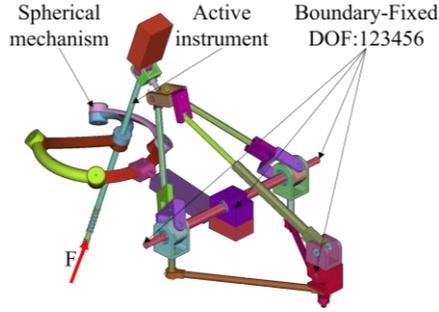
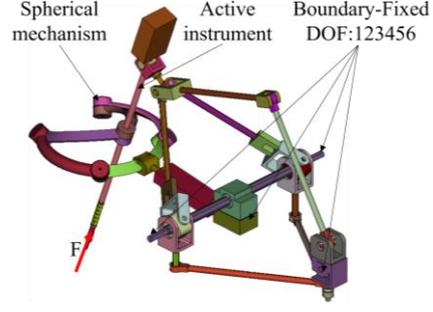

**Fig. 5a.** Load case scenario for ATHENA-1     **Fig. 5b.** Load case scenario for ATHENA-2

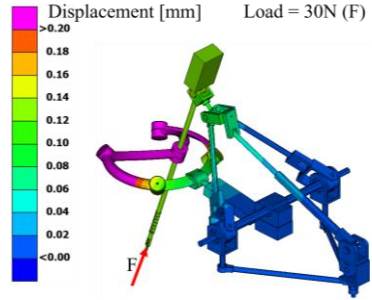
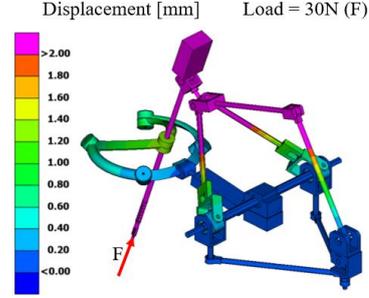

**Fig. 6a.** Overall displacement for ATHENA-1     **Fig. 6b.** Overall displacement for ATHENA-2

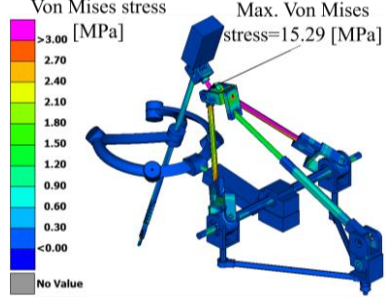
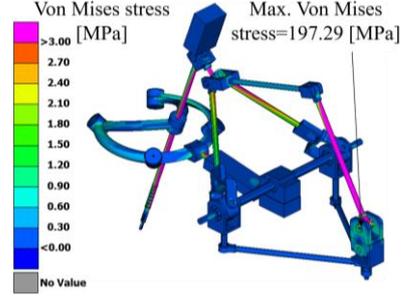

**Fig. 7a.** Von Mises stress for ATHENA-1     **Fig. 7b.** Von Mises stress for ATHENA-2

To determine the overall stiffness of the two robots, the displacement of the node where the force was applied (from the tip of the instrument) was read and the stiffness was calculated using Eq. 11.

$$K = F / \delta \qquad (11)$$

where F is magnitude of the applied force and the magnitude of displacement where the force was applied, the stiffness of the two robots was calculated.

For ATHENA-1, the displacement of the node where the force was applied is 0.23 mm while for ATHENA-2, the displacement of the same node is 3.96 mm. The calculated stiffness for ATHENA-1 parallel robot is 130.43 N/mm, while the calculated stiffness for ATHENA-2 is 7.58 N/mm. The calculated stiffness of the two robots (at the tip of the instrument) can be correlated and confirmed by the deformation behavior presented in Fig. 6a. and Fig. 6b. The much higher deformation of ATHENA-2 compared to the lower of ATHENA-1 is caused only by the revolute joints of the modified kinematic chain. This alters the decomposition of the reaction force vectors coming from the instrument, leading to higher deformation and reduced overall stiffness. For both parallel robots, the highest von Mises stress is significantly below the Yield stress limit of the corresponding materials showing that the components are not under-dimensioned, and the von Mises stress is generally lower in ATHENA-1 compared to ATHENA-2. In conclusion, the results of the two FEM simulations show that ATHENA-1 is structurally superior to ATHENA-2, with higher stiffness, lower deformations and lower von Mises stresses. The results are summarized in Table 1, where results for commercial robots in the field of minimally invasive surgery are also presented.

**Table 1. FEM simulation results compared to commercial robots**.

| Robot name | Displacement [mm] | Stiffness [N/mm] | von Mises stress [MPa] |
| --- | --- | --- | --- |
| ATHENA-1 | 0.23 | 130.43 | 15.29 |
| ATHENA-2 | 3.96 | 7.58 | 197.29 |
| Commercial robot | 0.2-0.5 | 5-10 | 30-65 |

## 4 Workspace analysis of the two parallel architectures

The workspace quantitative analysis of ATHENA-1 and ATHENA-2 is performed to study the usability of each architecture related to the medical task. Using Eqs. (1-11) and the variation intervals of $X_E \in [0, 300]$ mm, $Y_E \in [-500, 500]$ mm and $Z_E \in [-350, 0]$ mm with 2 mm increment, and the detailed data described in [22-23], the graphical representation of the two workspaces were performed using MATLAB. A series of constraints were used as follows: $q_1$, $q_2$ and $q_3 \in \mathbb{R}$, $l_{ins} < 250$ mm, $q_1 \in [0...l_1]$, $q_2 \in [0...2l_1]$. Additionally, for ATHENA-1 $q_3 > l_{2min}$, $q_3 < l_{2max}$, where $l_{2min}$ and $l_{2max}$ is the active stroke of the linear actuator, while for ATHENA-2 $q_3 \in [-45°, 45°]$. The graphical representations of the workspaces generated for the two architectures are illustrated in Fig. 8 and Fig.9.

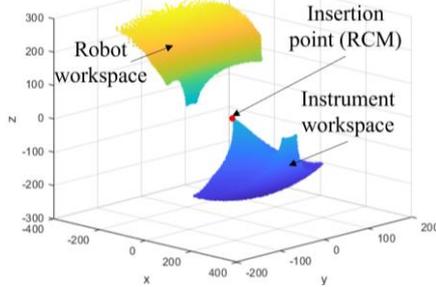
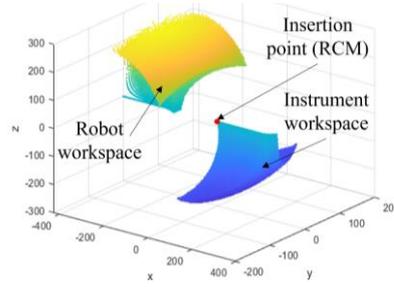

**Fig. 8.** ATHENA-1 workspace　　　　**Fig. 9.** ATHENA-2 workspace

A number of 196817 valid points have been generated for ATHENA-1 and 241586 for ATHENA-2. This means that ATHENA-2 has a workspace higher by approximately 22.7% than the parallel robot ATHENA-1, which was expected. Based on the analysis performed in [22], both ATHENA-1 and ATHENA-2 parallel robots cover the required workspace, which was defined at the training center of the University of Medicine and Pharmacy "Iuliu Hatieganu" in Cluj-Napoca, the methodology being detailed [22]. The workspace illustrated in Figs. 8 and 9 is a free singularity workspace for both parallel robots, the singularities analysis being presented in [22].

## 5    Conclusions

This paper presents the kinematic schemes, 3D designs, FEM simulation, and workspace analysis of two 4-DOF parallel architectures (ATHENA-1 and ATHENA-2) designed for minimally invasive pancreatic surgery. The FEM simulation presents the validation of the mechanical design of the two 3D models whit the elements being dimensioned well below the yield strength of the materials used. The stiffness analysis shows that the ATHENA-1 is much stiffer (130.43 N/mm) than the ATHENA-2 (7.58 N/mm), both being studied in similar configurations. Based on this stiffness analysis, the ATHENA-1 is selected for developing the experimental model. In addition, a comparative analysis based on their workspace was also carried out, the results showing a higher workspace for ATHENA-2, while both workspaces cover the required volume of the surgical task. Future developments will focus on the development of the command and control logic of ATHENA-1, as well as on the development of the experimental model and experimental testing under laboratory conditions.

## Acknowledgements

This research was funded by the project New smart and adaptive robotics solutions for personalized minimally invasive surgery in cancer treatment - ATHENA, funded by European Union – Next Generation EU and Romanian Government, under National Recovery and Resilience Plan for Romania, contract no. 760072/23.05.2023, code CF 116/15.11.2022, through the Romanian Ministry of Research, Innovation and Digitalization, within Component 9, investment I8.